\documentclass[12pt]{article}
\usepackage{amsmath}
\usepackage{graphicx}
\usepackage{float}
\usepackage{enumerate}
\usepackage{booktabs}
\usepackage{hyperref}
\usepackage{rotating}
\usepackage{caption}
\usepackage[]{algorithm2e}

\usepackage{natbib}
\date{July 16, 2024}

\usepackage{enumitem}
\usepackage{url} % not crucial - just used below for the URL 

\newcommand{\blind}{1}

\hypersetup{
    colorlinks=true,
    linkcolor=blue,      % Color for internal links (e.g., table of contents)
    citecolor=blue,     % Color for citation links (e.g., in-text references)
    urlcolor=blue,    % Color for URLs
}

\begin{document}

\def\spacingset#1{\renewcommand{\baselinestretch}%
{#1}\small\normalsize} \spacingset{1}

%%%%%%%%%%%%%%%%%%%%%%%%%%%%%%%%%%%%%%%%%%%%%%%%%%%%%%%%%%%%%%%%%%%%%%%%%%%%%%

\if1\blind
{
  \title{\bf Bayesian Causal Forests for Longitudinal Data: Assessing the Impact of Part-Time Work on Growth in High School Mathematics Achievement}
  \author{\hspace{-1.5cm}Nathan McJames$^{1,2,}$\thanks{
    Corresponding Author: nathan.mcjames.2016@mumail.ie.\\
    This work has emanated from research conducted with the financial support of Science Foundation Ireland under grant number 18/CRT/6049. In addition Andrew Parnell's work was supported by: a Science Foundation Ireland Career Development Award (17/CDA/4695) and SFI Research Centre award (12/RC/2289\_P2). For the purpose of Open Access, the authors have applied a CC BY public copyright licence to any Author Accepted Manuscript version arising from this submission.} , Ann O'Shea$^{2}$, Andrew Parnell$^{1,2}$\hspace{.2cm}\\
    \hspace{-1.5cm}$^{1}$Hamilton Institute, Maynooth University, Co. Kildare, Ireland\\
    \hspace{-1.5cm}$^{2}$Department of Mathematics and Statistics, Maynooth University, Co. Kildare, Ireland\\
    }
  \maketitle
} \fi

\if0\blind
{
  \bigskip
  \bigskip
  \bigskip
  \begin{center}
    {\LARGE\bf Bayesian Causal Forests for Longitudinal Data: Assessing the Impact of Part-Time Work on Growth in High School Mathematics Achievement}
\end{center}
  \medskip
} \fi

\bigskip
\begin{abstract}
Modelling growth in student achievement is a significant challenge in the field of education. Understanding how interventions or experiences such as part-time work can influence this growth is also important. Traditional methods like difference-in-differences are effective for estimating causal effects from longitudinal data. Meanwhile, Bayesian non-parametric methods have recently become popular for estimating causal effects from single time point observational studies. However, there remains a scarcity of methods capable of combining the strengths of these two approaches to flexibly estimate heterogeneous causal effects from longitudinal data. Motivated by two waves of data from the High School Longitudinal Study, the NCES' most recent longitudinal study which tracks a representative sample of over 20,000 students in the US, our study introduces a longitudinal extension of Bayesian Causal Forests. This model allows for the flexible identification of both individual growth in mathematical ability and the effects of participation in part-time work. Simulation studies demonstrate the predictive performance and reliable uncertainty quantification of the proposed model. Results reveal the negative impact of part time work for most students, but hint at potential benefits for those students with an initially low sense of school belonging. Clear signs of a widening achievement gap between students with high and low academic achievement are also identified. Potential policy implications are discussed, along with promising areas for future research.
\end{abstract}

\noindent%
\vfill
\newpage
\spacingset{1.9} % DON'T change the spacing!
\spacingset{1} % DON'T change the spacing!
\section{Introduction}

For many high school students, part-time jobs have become an integral part of their daily routine, just as important as homework, studying, and completing assignments \citep{singh2000effect}. The reasons for seeking part-time work can vary widely among students. Some work to support their families financially, others to develop their character, gain maturity, or simply to earn spending money \citep{kablaoui1991effects}. Regardless of the reasons for students choosing to work part-time, however, this work can have a significant impact on their educational journey \citep{bachman2014part}. Our study introduces a new approach for modelling individual level growth in student achievement, and explores the causal effect of intensive part-time work on this growth, where part-time work is defined as upwards of 20 hours of work per week during the school year \citep{lee2007work}.

Estimating causal effects from longitudinal data is a challenging but essential task. Established methods include inverse probability weighting \citep{hogan2004instrumental}, two-way fixed effects \citep{imai2021use}, and difference-in-differences \citep[DiD,][]{donald2007inference}. A key limitation of many of these approaches is that they often rely on strong assumptions that may not be appropriate for the target data. The parallel trend assumption of the difference-in-differences method, for example, assumes that the treatment group would have followed a similar trajectory to the control group had they not received treatment \citep{roth2023s}. This can easily be violated in practice, as confounding variables may influence both the probability of receiving treatment and the trajectories in the outcome of interest. Students who self select into part-time work, for example, may experience less growth than their peers even without part-time work \citep{monahan2011revisiting}. Some work has been conducted to tackle this limitation by relaxing the assumption of parallel trends conditional on covariates \citep{abadie2005semiparametric, callaway2021difference}, but important limitations remain.

Other methods rooted in structural equation modelling such as G-estimation \citep{robins1997causal}, and longitudinal extensions of targeted minimum loss based estimation \citep{lendle2017ltmle} excel in estimating causal effects from longitudinal data when faced with challenges such as drop-out, time varying covariates, and dynamic treatment regimes. A weakness of these methods, however, is that they are often restricted to estimating average causal effects, without the ability to explore individual level variations or heterogeneity in responses to treatment. This is an important limitation, especially in the context of part-time work, as there is research to show that the effects of part-time employment can vary significantly depending on factors such as gender, motivations for working part-time, and socioeconomic backgrounds \citep{entwisle2000early}.

When understanding heterogeneity in causal effects is important, Bayesian non-parametric methods based on Bayesian Additive Regression Trees \citep{hill2011bayesian, chipman2010bart} and Bayesian Causal Forests \citep{hahn2020bayesian} have become the gold standard. The default implementations of these methods are only applicable to single time point observational data, however, precluding the study of trends in educational outcomes over time. Our study extends BART and BCF to the setting of longitudinal data. By combining the flexibility of these methods with the highly interpretable structure of the difference-in-differences model, we simultaneously relax the parallel trend assumption of the DiD methods, while also allowing for the study of individual level variations in the growth curves of student achievement, and the heterogeneous impact of part-time work on this growth.

While other studies \citep{wang2024longbet} have also introduced longitudinal extensions of BART and BCF, with a focus on situations where there is a staggered adoption of treatment, our proposed model assumes a very different structure, and includes three important features targeted specifically at our motivating data. First, our model places separate priors directly over the growth trajectories and the effects of treatment on this growth. This allows us to inform the model with prior information, and comes with the added advantage of allowing the incorporation of model explainability tools, and variable importance metrics directly associated with the parameters of interest \citep{inglis2022visualizations, inglis2022visualizing}. The model structure also accommodates time varying covariates, such as evolving levels of student motivation which are common in education studies. Finally, while not the main focus of the present study, an additional feature not included in BCF models before is the ability to handle missing data in the covariates or the treatment assignment. We tackle this issue with a feature borrowed from \citet{kapelner2015prediction}, and a novel update step for the treatment status indicator.

The remainder of our paper is structured as follows: In Section~\ref{data section} we describe our motivating dataset, the High School Longitudinal Study of 2009 (HSLS), and outline some key features of the data. Section~\ref{methodology section} introduces the proposed model, and shows how we extend BART and BCF to provide a foundation for estimating growth curves of student achievement and heterogeneous treatment effects of part-time work. To further support the credibility of our proposed methodology, Section~\ref{sim section} applies our model to simulated data designed to mirror the characteristics of the HSLS dataset. We benchmark our performance against other potential candidate models, showcasing the unique capabilities of our model in overcoming challenges that remain difficult for existing approaches. In Section~\ref{application section}, we deploy our model to the HSLS data and present the results of our study. Finally, we conclude our paper in Section~\ref{discussion} with a discussion of our findings, implications for policy, and areas for future work.

\section{Data}
\label{data section}

The High School Longitudinal Study of 2009 \citep{ingels2011high} is an ongoing study of a nationally representative sample of high school students in the US. It is the most recent of a series of five longitudinal studies launched by the National Center for Education Statistics. The first wave of data collection for HSLS took place in the fall of 2009 at the start of the academic year when the students were in the ninth grade. More than 20,000 high school students took part in this part of the study. A follow up of these students then took place in the spring of 2012 when the students were in the eleventh grade. Further follow ups have also taken place in 2013, 2016, and 2017 to discover how the students are progressing in the years after high school, but did not involve mathematics achievement tests so we will focus solely on the first two waves of the data. Due to some students dropping out of high school, some schools closing, and others disagreeing to continue their participation in the study, the second wave of data collection involved just under 19,000 of the original Wave 1 sample.

Data collection during HSLS followed a similar procedure during both waves of the study. Mathematics achievement was assessed during both waves using a computer delivered assessment with questions designed to measure the algebraic reasoning abilities of the students. The resulting achievement estimates assigned to the students were calculated using Item Response Theory \citep{cai2016item}. The contextual data gathered as part of the study was based on a survey answered by the students, a parent, school administrators, and school teachers. Information collected from the student survey includes characteristics such as sex, age, self concept in mathematics, sense of school belonging, and other details such as participation in activities like part-time work. Data from the parent survey includes important socioeconomic variables such as family income, parental employment and education. School related data includes information such as the administrator's perception of the overall climate within the school, and the level of expectations of student academic success.

To ensure a representative sample of students, a stratified, two-stage random sampling design was employed by the study organisers. This involved first approaching eligible high schools, 944 of which agreed to participate in the study, and then randomly sampling students from each of those schools, leading to a total sample of 21,444 participating students. Sampling weights resulting from this design are provided in the dataset to account for non-participation bias and were used to appropriately weight the results discussed later in the paper. Table~\ref{side_table} of the supplementary material provides weighted summary statistics for a subset of the categorical variables from the base year (Wave 1), and also provides mean achievement levels from both waves of the study.

Our study uses the public-use version of the HSLS data. Some of the data in this public use version of the dataset has been obfuscated or removed in order to maintain the anonymity of the students and the schools who took part in the study. Therefore, a school identifier indicating which students attend the same school is not available in this version of the dataset, precluding a hierarchical modelling strategy. The restricted use version of the dataset does include this information but is only available with strict controls in place. This is a limitation of our study, but ensures our results are more easily reproducible without requiring a restricted use version of the dataset. Furthermore, there is evidence to suggest that part-time work is more likely to be influenced by student and family related variables than school related variables \citep{howieson2012new}, partially mitigating the potential for unmeasured confounders to bias our results.

\section{Methodology}
\label{methodology section}

\subsection{The Model}

Our motivating dataset consists of two waves, but for the sake of generality in this section we will describe how the model applies to datasets of up to $T$ waves of student data. We are interested in modelling trajectories of student achievement where we have data on $n_{1}$ students participating in an initial base year assessment, and subsequent follow-ups on $n_{2}\ldots n_{T}$ of the same students during waves $2$ to $T$. We allow for the possibility of drop-out, whereby $n_{T} \leq n_{T-1} \leq \ldots \leq n_{1}$. We will represent the contextual data associated with student $i$ up to time $t$ by $x_{i,t}$, where $t=1$ indicates the data is from the base year (Wave 1), and subsequent values of $t$ indicate the data encompasses extra information collected up to and including Wave $t$. We will not distinguish between data from different surveys or questionnaires, so $x_{i,t}$ captures all of the student, parent, and school level data associated with student $i$ up to time $t$. Given the accumulation of information on students over time as they complete more surveys from additional waves, the number of columns in $x_{i,t}$ will be less than the number of columns in $x_{i,t+1}$. To distinguish between students who do and do not work part-time, we will let $Z_{i, t+1}$ be a binary indicator of length $n_{t+1}$ which indicates for each student if they reported having a part-time job which involved them working on average 20 hours or more per week during the period between Waves $t$ and $t+1$. For the achievement data, let $y_{i,t}$ denote the observed mathematics achievement of student $i$ recorded at time $t$.

Our research questions concern two quantities of interest. The first is related to the growth in mathematics achievement between Waves $t$ and $t+1$, which we will denote by $G_{i,t+1}=y_{i,t+1}-y_{i,t}$. The second concerns the impact of part-time work on this growth. To understand this impact, we adopt the Neyman-Rubin causal model \citep{splawa1990application, sekhon2008neyman}, and postulate that for each individual $i$, there are two potential growth values. One that would be observed if the student worked part-time, $G_{i,t+1}(Z_{i,t+1}=1)$, and one that would be observed if the student did not, $G_{i,t+1}(Z_{i,t+1}=0)$. With these quantities defined, the impact of part-time work on the growth in student achievement during this period is captured by $\tau_{i,t+1}=G_{i,t+1}(Z_{i,t+1}=1)-G_{i,t+1}(Z_{i,t+1}=0)$. Of course, we only ever observe one of these potential growth values, namely $G_{i,t+1}=G_{i,t+1}(Z_{i,t+1}=1)Z_{i,t+1}+G_{i,t+1}(Z_{i,t+1}=0)(1-Z_{i,t+1})$, so we make the following assumptions:
\begin{enumerate}[start=1,label={\bfseries Assumption \arabic*:}, left=0.5in]
    \item The Stable Unit Treatment Value Assumption. We assume that the potential growth values of every student $i$ between periods $t$ and $t+1$ are independent of whether or not any other student $j$ worked part-time in any period.
    \item The Sequential Ignorability Assumption. We assume that conditional on their observed characteristics and treatment history up to the period of interest, the potential growth values of student $i$ are independent of whether or not they worked part-time. Notationally, we assume that $G_{i,t+1}(Z_{i,t+1}=0), G_{i,t+1}(Z_{i,t+1}=1) \perp Z_{i,t+1} | x_{i,t+1}, Z_{i,t}$.
    \item The Overlap Assumption. We assume that for every observed covariate and treatment history, there is a non-zero probability of working, or not working part-time during any period of interest: $0<P(Z_{i,t+1}=1|x_{i,t+1}, Z_{i,t})<1$. 
\end{enumerate}
\noindent If these conditions hold \citep{angrist1996identification, kurz2022augmented, myint2024controlling, hernan2020causal}, then we may write that
\[E[G_{i,t+1}(Z_{i,t+1})|x_{i,t+1}] = E[G_{i,t+1}|Z_{i,t+1}, x_{t+1}].\]
\noindent Our model for student achievement across all waves of data then becomes:
\[y_{i,t} = \mu(x_{i,1}) + \sum_{w=1}^{T-1}{\left(\underbrace{\delta_{w+1}(x_{i,w+1}, y_{i,1}\ldots y_{i,w}, \hat{\pi}_{i,w+1}) + \tau_{w+1}(x_{i,w+1}, y_{i,1}\ldots y_{i,w})Z_{i,w+1}}_{G_{w+1}(x_{i,w+1}, y_{i,1}\ldots y_{i,w}, \hat{\pi}_{i, w+1})}\right)I(t>w)} + \epsilon_{i,t}\]
where the different parts of the model work together in a cumulative fashion to predict different parts of a student's mathematics achievement. Predictions for achievement at Wave 1 are given by $\mu()$, while achievement at any subsequent Wave $t$ is given by adding this to a cumulative sum of achievement growths, $G_{w+1}()$. Within each time period, $\delta_{w+1}()$ and $\tau_{w+1}()$ represent the growth that would have been realised without part-time work, and the expected impact of part-time work on this growth respectively. The additional covariate $\hat{\pi}_{i,w+1}$ included in the $\delta_{w+1}()$ part of the model is a propensity score, which estimates the probability of observation $i$ receiving treatment during this period conditional on their covariates. This inclusion follows the advice of \cite{hahn2020bayesian}, who demonstrated that incorporating this ``clever covariate'' can help mitigate the issue of regularisation-induced confounding. Finally, $\epsilon_{i,t}$ represents the error term for student $i$ at time $t$, which we assume to be normally distributed with mean 0 and variance $\sigma^2$, $\epsilon_{i,t} \sim N(0, \sigma^2)$.

In our model, the contributions made by $\mu()$ and each of $\delta_{w+1}()$ and $\tau_{w+1}()$ come from ensembles of $n_{\mu}$, $n_{\delta}$, and $n_{\tau}$ regression trees based on the BART model of \citet{chipman2010bart}. For ease of exposition as we discuss the Bayesian backfitting MCMC algorithm by which the regression trees fit to the data, let us consider the simplest scenario where $n_{\mu}=n_{\delta}=n_{\tau}=1$ and $T=2$, leaving the general case for the supplementary material. The MCMC sampler begins with each of $\mu()$, $\delta_{2}()$, and $\tau_{2}()$ initialised as stumps (decision trees where the root is also the sole terminal node, and the terminal node parameter of each tree is set to zero). Next, each iteration starts by selecting at random one of four possible operations (grow, prune, change, or swap) to apply to the $\mu()$ tree in order to propose a new tree structure. This proposal is then accepted or rejected with a Metropolis-Hastings step before the terminal node (or now possibly nodes) of the $\mu()$ tree are updated via a Gibbs-sampling step which attempts to explain any leftover variation in the partial residual $y_{i,t}$ less the contribution from $\delta_{2}()$ and $\tau_{2}()$. Analogous operations are then applied to the $\delta_{2}()$ and $\tau_{2}()$ trees before the residual variance parameter is also updated via Gibbs-sampling. This cycle repeats for a specified number of iterations, providing a desired number of posterior draws for the tree structure and terminal node parameters of $\mu()$, $\delta_{2}()$, and $\tau_{2}()$, as well as the residual variance parameter $\sigma^2$.

Overfitting is prevented through the use of the tree prior from \citet{chipman2010bart} which specifies that the probability of any node at depth $d$ being non-terminal is given by $\alpha(1+d)^{-\beta}$. Therefore, for a tree $T$ with terminal nodes $h_{1} ... h_{K}$, and non-terminal nodes $b_{1} ... b_{L}$, we have that:
\[P(T)=\prod_{k=1}^{K} \alpha(1+d(h_{k}))^{-\beta} \prod_{l=1}^{L} [1-\alpha(1+d(b_{l}))^{-\beta}]\]
The strength of this prior can be adjusted through setting different values for $\alpha$ and $\beta$. For the $\mu()$ and $\delta()$ trees we adopt the default prior from \citet{chipman2010bart}, of $\alpha=0.95$, $\beta=2$, while for the $\tau()$ trees we impose stronger regularisation as we expect there to be less heterogeneity in the effects of part-time work than in $y$ itself, choosing $\alpha=0.25$, $\beta=3$ as suggested by \citet{hahn2020bayesian}.

To ensure each tree contributes approximately equally to the overall prediction, the terminal node parameters of each tree are given a normal prior. In each type of tree, we have
\[\mu \sim N(0, \sigma_{\mu}^2),\ \delta \sim N(0, \sigma_{\delta}^2),\ \tau \sim N(0, \sigma_{\tau}^2)\]
After scaling $y$ to follow a standard normal distribution prior to fitting the model, a sensible choice for $\sigma_{\mu}^2$ is $1/n_{\mu}$, ensuring the terminal mode parameters in the $\mu()$ trees have adequate room to cover the range of the data. Similarly, we use a prior of $\sigma_{\delta}^2=1/n_{\delta}$, but given we expect the magnitude of the treatment effects to be relatively small in comparison to $y$ we set $\sigma_{\tau}^2=0.5^2/n_{\tau}$. Finally, the conjugate prior for $\sigma^2$ is an inverse gamma distribution: 
\[\sigma^2 \sim \text{Inverse-Gamma}(\nu/2, \nu\lambda/2),\]
for which we have found a reliable default choice is to set $\nu=3$, and $\lambda=0.1$.

\subsection{Special Features}

Two challenges related to missing data required us to build some extra functionality into our model. The first challenge was related to missing data in the covariates. Missing data in the covariates can arise for several reasons in the dataset. For example, some questions may have been purposely skipped by students, their parents, or teachers, and other times the answer to a particular question may not have been known. On average, 1.9\% of the data was missing, and the most data missing for any particular variable was 19\%. Common approaches for dealing with missing data in the covariates include single or multiple imputation \citep{lin2020missing}, but an extra possibility specific to tree based models is the approach developed by \citet{kapelner2015prediction}, which involves treating missing data as an important feature of the data, operating under the assumption that the data is missing at random. To summarise, this approach involves directing observations with missing data to the left or right child of a node that is being split on, allowing the model to learn from any relationship between missingness and the outcome variable, handling missing data as an integral part of the model, thus accounting for uncertainty that may be present. 

The second challenge was related to missing data in the treatment $Z_{i,2}$ itself, as not all students answered the question on how many hours they worked part-time during school weeks. This type of missingness affected 3.4\% of the observations in the data. This challenge is addressed by introducing an additional Gibbs-sampling step at the end of each iteration of the MCMC sampler, where the missing $Z_{i,2}$ values are themselves treated as parameters to be updated, with prior probability $p_{i}$, conditional on the rest of the data: 
\[P(Z_{i,2}=1|...)=\frac{1}{1+\left(\frac{1-p_{i}}{p_{i}}\right)\left(e^{\frac{1}{2\sigma^2}\left[(y_{i,2}-\mu_{i}-\delta_{i}-\tau_{i})^{2}-(y_{i,2}-\mu_{i}-\delta_{i})^{2}\right]}\right)}.\]
\noindent These two added features allowed us to keep a full representative sample of students while accounting for the added uncertainty introduced into our results by the presence of missing data.

One final challenge that is common when working with assessment data of student achievement is the use of plausible values \citep{wu2005role, khorramdel2020plausible}. In order to prevent the computer delivered assessment from taking unduly long, it was only possible for HSLS to present each student with a limited number of questions. This introduces some room for error in the achievement estimates of the students, and as a result, HSLS provides researchers with five plausible values of student achievement from the posterior of each student's achievement estimate. In line with best practice, we therefore ran five chains of our model, one applied to each plausible value of student achievement, and pooled them together after burn-in to appropriately handle this uncertainty.

\subsection{Alternative Methodologies}

Our work shares connections with several areas of Bayesian non-parametric modelling, and longitudinal methods for causal inference. First there is the clear connection to BART \citep{chipman2010bart}, as this method provides a foundation for the different parts of our model. BART based methods have become popular in the area of causal inference \citep{hill2011bayesian} and have demonstrated impressive performance and reliable uncertainty quantification. A second strong connection is with the Bayesian Causal Forest model developed by \citet{hahn2020bayesian}, which also uses BART as a foundation for estimating causal effects. Both methods could potentially be applied to our research questions but fall short of offering the same abilities in this context as our own model in several important ways which are worth discussing.

The most natural way for BCF to be applied to our problem setting would be to manually calculate the growth values $G_{i,t+1}$ for each student $i$, and each time period $t$ to $t+1$. Applying the BCF model to a specific time period would then yield the following, allowing us to recover what our model captures with the $\delta_{t+1}()$ and $\tau_{t+1}()$ part of the model:
\[G_{i,t+1}=\delta_{t+1}(x_{i,t+1}, y_{i,t}, \hat{\pi}_{i,t+1}) + \tau_{t+1}(x_{i,t+1}, y_{i,t})Z_{i,t+1}+\epsilon_{i},\ \epsilon_{i}\sim N(0, \sigma^2)\]
A key limitation of this approach is that it does not model the full data generating process, only the growth between Waves $t$ and $t+1$. This means that students who participate in Wave $t$ but not in Wave $t+1$ (and consequently have no calculable $G_{i,t+1}$) are excluded from the model and are unable to inform the predictions made by the model. Secondly,  manually calculating the $G_{i,t+1}$ growth values (to be used as the response variable in this approach), is likely to lead to a smaller signal to noise ratio in the response as the error terms from $y_{i,t}$ and $y_{i,t+1}$ combine, making it more difficult for the model to detect the relationships it is trying to model. A BART only approach could also be applied to the data in a similar way using $G_{i,t+1}$ as the response, but this approach would share the same limitations. Of course, if treatment effects were the only quantity of interest then it would also be possible to apply BART or BCF directly to $y_{i,t+1}$, but this would preclude any inference on the growth values $G_{i,t+1}$, so would fail to address this aspect of our study.

Researchers more familiar with difference-in-differences \citep{roth2023s} based approaches might like to think of our model as a Bayesian non-parametric DiD model where our $\delta_{t+1}()$ trees model the difference for the control group (non part-time workers), and our $\tau_{t+1}()$ trees model the difference in this difference experienced by the treatment group (the part-time workers). Crucially, our approach handles this situation much more flexibly than traditional DiD based methods, as the flexibility of the $\delta_{t+1}()$ trees means we can relax the assumption of parallel trends conditional on the covariates of the students, and the $\tau_{t+1}()$ trees also allow us to capture heterogeneity in the effects of part-time work which is often not possible with DiD based methods. See the supplementary material for an illustration of how our proposed model fits into this framework.

Finally, our method also shares similarities with causal methods applicable to longitudinal data such as G-estimation \citep{tompsett2022gesttools}, or longitudinal extensions of targeted minimum loss based estimation \citep[LTMLE,][]{lendle2017ltmle}. Both methods have gained popularity owing to their ability to handle complex situations such as time varying confounding, situations where the primary interest is in the causal effect of a series of sustained or irregular treatments, and where the interest is in the lagged effects of a treatment. Our focus however, will be on heterogeneity in the direct effect of a single period of part-time work on the immediately following mathematics assessment, which is not achievable with the available implementations of these methods. Additionally, our model will also provide insights into the growth trajectories of student achievement, a feature that is not modelled by these other approaches.

\section{Simulation Studies}
\label{sim section}

In this section, we assess our proposed model's performance in a simulation study designed to match the features of the motivating HSLS data. We also compare our proposed model with alternative approaches in order to highlight the added performance offered by our method. Our simulation study consists of two data generating processes. DGP1 focuses on heterogeneity in treatment effects and growth curves, making it well-suited to flexible approaches based on BART and BCF. It features two waves of data to accommodate the alternative methods which can not handle multiple time periods. DGP2 is inspired by a synthetic dataset from the R package \texttt{gesttools}. This process includes more than two time points and features time-varying covariates. It focuses on estimating the Average Treatment Effect, enabling a fair comparison of our method with the \texttt{gesttools} and LTMLE packages, which do not support the estimation of heterogeneous treatment effects, but are correctly specified for the features of this DGP.

\subsection{Data Generating Process 1}

Our first data generating process is based on a modified version of the first Friedman dataset \citep{friedman1991multivariate}, a common benchmarking dataset featuring non linear effects and interaction terms. We will use this dataset to assess how well each of the flexible causal machine learning methods can capture heterogeneity in the growth curves of student achievement, and the treatment effects themselves. We simulate ten covariates measured at Wave 1: $x_{1} \ldots x_{10}$, and a second observation of each of these ten variables again at the final Wave 2: $x_{11} \ldots x_{20}$, where the second observation of each variable is equal to the first plus a small amount of random noise, e.g., $x_{16}=x_{6}+r$, with $r$ a uniform random variable between 0 and 0.4. The structure of the simulated achievement level of each student is of the form described earlier:
\[y_{i,t} = \mu(x_{i,1}) + \underbrace{\delta_{2}(x_{i,2}, y_{i,1}, \hat{\pi}_{i,2})I(t>1) + \tau_{2}(x_{i,2}, y_{i,1})Z_{i,t+1}I(t>1)}_{G(x_{i,2}, y_{i,1}, \hat{\pi}_{i,2})} + \epsilon_{i,t},\ \epsilon_{i,t}\sim N(0, \sigma^2)\]
where $\mu(x_{i,1}) = 10\sin(\pi x_{1}x_{2})+20(x_{3}-0.5)^2+10x_{4}+5x_{5}$, $\delta_{2}()=\dfrac{1}{3}\mu(x_{i,1})+3x_{11}^2+2x_{15}^2$, and $\tau_{2}()=-x_{4}-x_{14}^2-x_{15}^3$. The true propensity scores are given by $p_{i}=P(Z_{i,2}=1|x_{i,2})=\text{Plogis}(\mu_{i}^{*}+\delta_{i}^{*})$, where $\mu_{i}^{*}+\delta_{i}^{*}$ is a normally scaled version of each of the original $\mu_{i}+\delta_{i}$ values. 

The compared methods are our longitudinal BCF model, BART using the approach outlined in \citet{hill2011bayesian}, a standard BCF model from \citet{hahn2020bayesian}, and the causal Generalised Random Forest model (GRF) from \citet{wager2018estimation}. The recently proposed BCF extension by \citet{wang2024longbet} would also make an excellent method for comparison when a documented R package becomes available. 

As outlined earlier, given that the longitudinal BCF model is the only one capable of directly modelling the growth curves, we will apply the other competing methods to the transformed outcome $y_{i,2}-y_{i,1}$, the difference in outcomes between Waves 1 and 2, to enable the prediction of growth using BART and BCF. For the longitudinal BCF model, we use 100 trees in the $\mu()$ part of the model responsible for predicting $y_{i,1}$ at Wave 1, 70 trees in the $\delta()$ part of the model responsible for predicting the growth under control, and 30 trees in the $\tau()$ part of the model responsible for predicting the heterogeneous treatment effects. For the standard BCF approach, we use 170 trees in the prognostic part of the model which will provide estimates for the growth under control, and 30 trees for estimating the treatment effects. The BART and GRF approaches both use 200 trees in total. Each simulation consists of 500 training observations, and 1000 test observations. The Bayesian methods are run for 500 burn-in and 500 post burn-in iterations. Satisfactory convergence was assessed via visual inspection of the posterior samples for a small subset of the 1000 replications of the data generating process.

\begin{figure}
    \centering
    \includegraphics[width=15cm]{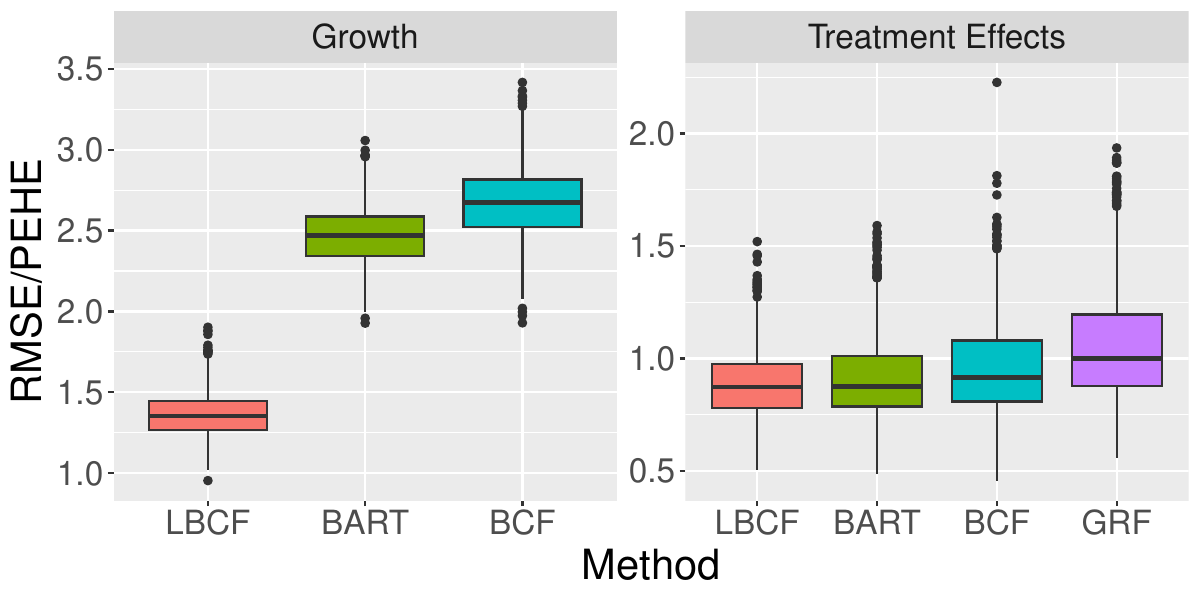}
    \caption{Visualisation of RMSE and PEHE metrics evaluated over 1000 replications of DGP1 for the BART, BCF, GRF, and LBCF models. In the left panel, which displays the RMSE of the $\delta_{i}$ predictions, the LBCF approach is clearly the strongest performer, with considerably lower RMSE values. In the right panel, which visualises the PEHE metrics, LBCF is again the strongest performer, but by a narrower margin.}
    \label{dgp1_plot}
\end{figure}

Table~\ref{dgp1_table} summarises how the compared approaches perform when tasked with predicting $\delta_{i}$ and $\tau_{i}$ across 1000 replications of the simulation. For $\delta_{i}$, this performance is evaluated using the average root mean squared error (RMSE) over the 1000 simulations: $RMSE=\sqrt{\frac{1}{N}\sum_{i=1}^{N}(\delta_{i}-\hat{\delta_{i}})^2}$. The equivalent metric used for $\tau_{i}$ is the precision in estimating heterogeneous effects (PEHE): $PEHE=\sqrt{\frac{1}{N}\sum_{i=1}^{N}(\tau_{i}-\hat{\tau_{i}})^2}$, also averaged over the 1000 simulations. Mean coverage rates of the 95\% credible intervals, bias, and credible interval widths are also provided for both $\delta_{i}$ and $\tau_{i}$. A visual representation of these results can be found in Figure~\ref{dgp1_plot}.

The clearest differences in Figure~\ref{dgp1_plot} relate to model performance predicting the $\delta_{i}$ values, with our proposed LBCF model achieving much lower RMSE values. Comparison with the GRF model was not possible here, as the GRF model output only provides treatment effect estimates. In the right panel of Figure~\ref{dgp1_plot}, the differences are more subtle, but the proposed model performs marginally better than the BART and BCF methods, which in turn both outperform the GRF based approach.

\begin{table}[h]
    \centering
    %\begin{tabular}{p{3cm}p{1.2cm}p{1.2cm}p{1.2cm}p{1.2cm}p{1.2cm}p{1.2cm}p{1.2cm}p{1.2cm}}
    \begin{tabular}{p{4cm}r r r r r r r}
        \hline
        \multicolumn{1}{c}{} & \multicolumn{3}{l}{$\delta_{i}$ predictions} & \multicolumn{4}{l}{$\tau_{i}$ predictions} \\
        \cmidrule(lr){1-4} \cmidrule(lr){5-8}
        & LBCF & BART & BCF & LBCF & BART & BCF & GRF \\
        \cmidrule(lr){1-4} \cmidrule(lr){5-8}
        RMSE/PEHE & \textbf{1.362} & 2.470 & 2.671 & \textbf{0.886} & 0.907 & 0.958 & 1.057 \\
        Mean Absolute Bias & \textbf{0.265} & 0.311 & 0.303 & \textbf{0.324} & 0.424 & 0.424 & 0.604 \\
        95\% Coverage & \textbf{0.980} & 0.996 & 0.891 & \textbf{0.935} & 0.921 & 0.911 & 0.769 \\
        95\% CI Width & 6.559 & 14.306 & 8.624 & 3.409 & 3.343 & 3.515 & 2.654 \\
        \hline
    \end{tabular}
    \caption{Summary of important metrics measured for $\delta_{i}$ and $\tau_{i}$ predictions, averaged over 1000 simulations of DGP1. The proposed LBCF model performs competitively, achieving a lower mean RMSE and PEHE than the alternative models. Bias, coverage, and credible interval widths are also close to ideal. Best results are highlighted in \textbf{bold} where a clear winner exists.}
    \label{dgp1_table}
\end{table}

Finally, the LBCF estimates are the least biased of all the compared methods, and are accompanied by close to ideal coverage rates. The credible interval widths from the LBCF estimator are similar to the competing methods when estimating the treatment effects, but are considerably narrower than the competing methods when estimating the growth values, offering a high degree of precision.

\subsection{Data Generating Process 2}

Our second data generating process comes from the R package \texttt{gesttools} \citep{tompsett2022gesttools}, which implements G-estimation for longitudinal data. Our focus here is on estimating the average treatment effect. As described in \citet{tompsett2022gesttools}, the dataset includes:

\begin{itemize}
    \item A baseline covariate $U\sim N(0,1)$
    \item Covariates $L_{t}\sim N(1+L_{t-1}+0.5A_{t-1}+U), t=1,2,3, A_{0}=0$
    \item Exposure $A_{t}\sim \text{Bin}(1, \text{expit}(1+0.1L_{t}+0.1A_{t-1})), t=1,2,3$
    %\item Censoring indicator $C_{t}\sim \text{Bin}(1, \text{expit}(-1+0.001\times L_{t-1}+0.001\times A_{t-1})), t=2,3,4$
    \item Time varying outcome $Y_{t}\sim N(1+A_{t}+\gamma_{t}A_{t-1}+\sum_{i=1}^{t}L_{t}+U, 1), t=2,3,4$
    \item Constants $(\gamma_{1}, \gamma_{2}, \gamma_{3})=(0, 0.5, 0.5)$
\end{itemize}

In this simulation study, the baseline covariate $U$ remains fixed, while the time varying covariates $L_{t}$ change at each wave in response to the values of the preceding covariates, and whether or not treatment was received. The likelihood of receiving treatment also depends on previous covariates and treatments. Note that while the time varying outcome depends on the treatment status at the current and previous time points, we will only estimate the direct effect of treatment at time $t$ on $y_{t}$. 

The methods we will compare are G-estimation as implemented by \texttt{gesttools}, longitudinal targeted minimum loss based estimation from the LTMLE package, and our proposed method. The \texttt{gesttools} and LTMLE approaches will use the default settings of the R packages, which make them the correctly specified models, while our approach will use the same setup from the previous simulation study. As before, we will run 1000 replications of the simulation study, but will evaluate performance on the training sample of 500 observations (the LTMLE and \texttt{gesttools} packages can not make predictions on unseen data).

\begin{figure}
    \centering
    \includegraphics[width=15cm]{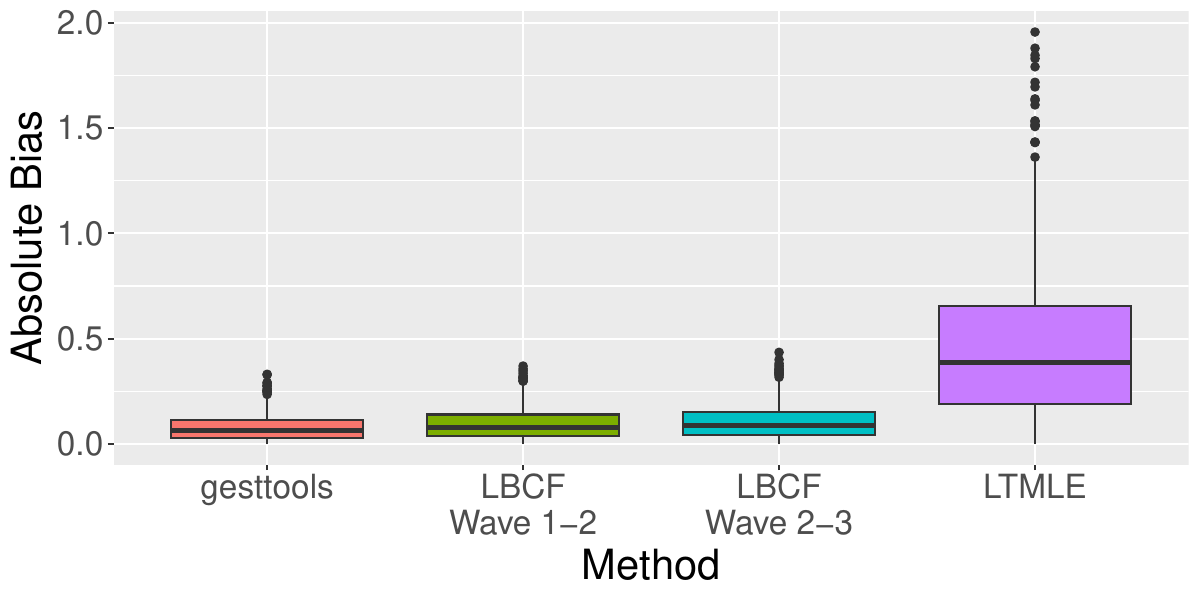}
    \caption{Visualisation of bias in ATE estimates over 1000 replications of DGP2 for the \texttt{gesttools}, LBCF, and LTMLE models. The \texttt{gesttools} package, which assumes a constant treatment effect at all time points shows minimal bias. This strong performance is closely followed by the proposed LBCF model, which provides estimates for the treatment applied between Waves 1 and 2, and 2 and 3. The LTMLE estimates appear to be much more biased.}
    \label{dgp2_plot}
\end{figure}

Figure~\ref{dgp2_plot} visualises the ATE estimates from the proposed approach, the \texttt{gesttools} package, and the LTMLE package. For the \texttt{gesttools} and LTMLE results, only one boxplot is shown. In the case of the \texttt{gesttools} results, this is because the package assumes the treatment has the same effect at all time points. In this simulation, this assumption is valid, but in general, the ability of our model to provide separate estimates at each time point is likely to be valuable. With the LTMLE package, it is necessary to define a contrast in order to estimate the effect of some sequence of treatments on the final observed outcome variable (in this case $Y_{3}$). For the simulation above, we tasked the LTMLE package with estimating the effect of the treatment sequence ($A_{1}=0, A_{2}=0, A_{3}=1$) relative to ($A_{1}=0, A_{2}=0, A_{3}=0$). This will recover the direct effect of $A_{3}$ on $Y_{3}$, which is equal to the direct effect of $A_{t}$ on $y_{t}$, consistent across time. In contrast, our proposed LBCF model is able to provide ATE estimates for both the effect of $A_{2}$ on $Y_{2}$, and the effect of $A_{3}$ on $Y_{3}$, offering a more detailed and flexible analysis.

Figure~\ref{dgp2_plot} visualises the absolute bias in estimating the ATE for each of the approaches over 1000 replications of DGP2. The \texttt{gesttools} package is the best performer here, closely followed by the LBCF estimates which are consistently accurate across both time periods. We note, however, that the \texttt{gesttools} package assumes the treatment effect is the same at all time points, and this is unlikely to always be valid. The bias of the LTMLE package is consistently much higher, indicating the model often struggled to identify the true ATE from the data.

A similar pattern is observed in Table~\ref{dgp2_table}, which provides additional information on the coverage rates, and mean credible/confidence interval widths. Here, the coverage achieved by the \texttt{gesttools} package and the two estimates provided by the proposed LBCF model are very close to ideal. The LTMLE package appears to underestimate the uncertainty in its estimates, however, and only achieves 76.8\% coverage. The LBCF model's credible interval widths at both time points are slightly wider than those of the \texttt{gesttools} package but remain significantly narrower than the LTMLE package's confidence interval widths.

\begin{table}
    \centering
    \begin{tabular}{lllll}
        \hline
        Model/Metric& \texttt{gesttools} & LBCF Wave 1-2 & LBCF Wave 2-3 & LTMLE \\
        \hline
        Mean Absolute Bias & \textbf{0.077} & 0.097 & 0.107 & 0.476 \\
        95\% Coverage & \textbf{0.950} & 0.913 & 0.936 & 0.838 \\
        Mean 95\% CI Width & \textbf{0.384} & 0.438 & 0.502 & 1.688 \\
        \hline
    \end{tabular}
    \caption{Absolute bias, coverage rates, and credible/confidence interval widths averaged over 1000 replications of DGP2. Coverage rates are very good for \texttt{gesttools} and LBCF, but less than ideal for LTMLE. The \texttt{gesttools} package provides the most precise estimates, with slightly narrower confidence interval widths than LBCF. Best results are highlighted in \textbf{bold} where a clear winner exists.}
    \label{dgp2_table}
\end{table}

In summary, the results from both data-generating processes in our simulation study underscore the proposed model's ability to provide flexible and accurate predictions, even when confronted with highly non-linear growth patterns or heterogeneity in treatment effects. The model achieved near-ideal coverage rates, exhibited minimal bias, and produced narrower credible intervals compared to other non-parametric causal models. In the second data-generating process, where the proposed model was benchmarked against a correctly specified G-estimation model, the LBCF model matched its strong performance, without making the same assumption that the treatment effect was consistent over time. Encouraged by the robust performance of our proposed model, we proceed to the next section, where we apply the longitudinal BCF method to the motivating HSLS dataset to assess the impact of part-time work on student achievement.

\section{Application to High School Longitudinal Study}
\label{application section}

Recall that HSLS includes two waves of data, with student achievement and other background characteristics measured at both time points. We are interested in understanding the amount by which the mathematics achievement of the students increases between these waves, how this growth depends on the characteristics of the students, the effect of part-time work on this growth, and how this effect is potentially moderated by other observed variables.

We apply our model to this dataset using the same model structure from the simulation study, with the same number of trees, but run a larger number of burn-in (3000) and post burn-in iterations (2000), to ensure satisfactory convergence. As described in the methodology, missing data is handled internally by the model, so there is no requirement for multiple imputation. The plausible values of student achievement are appropriately accounted for by pooling 5 separate chains, each of which were applied to one of the 5 sets of plausible values. Sampling weights are also accounted for by appropriately weighting the average treatment effect results displayed below.

Figure~\ref{growth_plot} shows the posterior distribution of the average growth, and a histogram of the individual $\delta_{i}$ estimates for each student present in Wave 2 of the dataset. The average growth is close to 0.63, and the majority of the growth estimates are positive, indicating that most students are expected to increase their mathematics achievement between Waves 1 and 2. Within the sample there is large variation, however, with some students predicted to increase their mathematics achievement by up to 2 units on the achievement scale, while for a small number of students, mathematics achievement is actually predicted to decrease by a small amount. For context, achievement at Wave 1 was normally distributed with a mean of approximately 0, and a standard deviation of approximately 1. Therefore, an increase in achievement by two units, or two standard deviations, is quite significant.

\begin{figure}
    \centering
    \includegraphics[width=15cm]{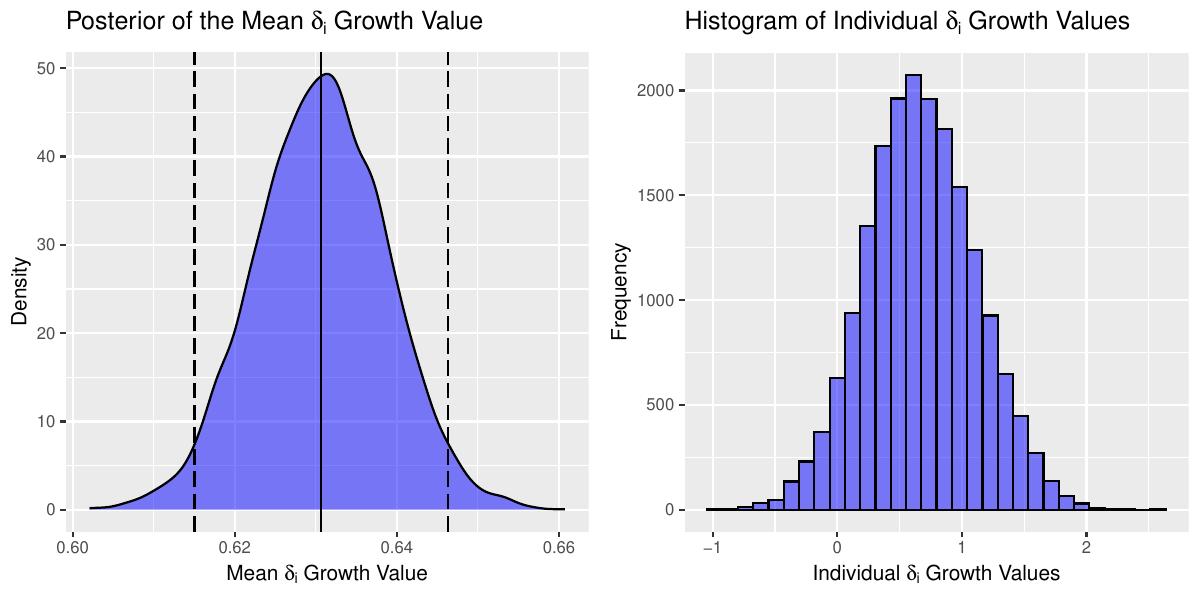}
    \caption{The left plot shows the posterior distribution of the average growth, while the one on the right displays a histogram of the individual $\delta_{i}$ estimates. The solid line in the left plot shows the posterior mean, while the dashed lines indicate a 95\% credible interval. Substantial variability is present in the $\delta_{i}$ values, indicating that some students are predicted to increase their achievement by much more than others who may even experience a decrease in achievement.}
    \label{growth_plot}
\end{figure}

\begin{figure}
    \centering
    \includegraphics[width=15cm]{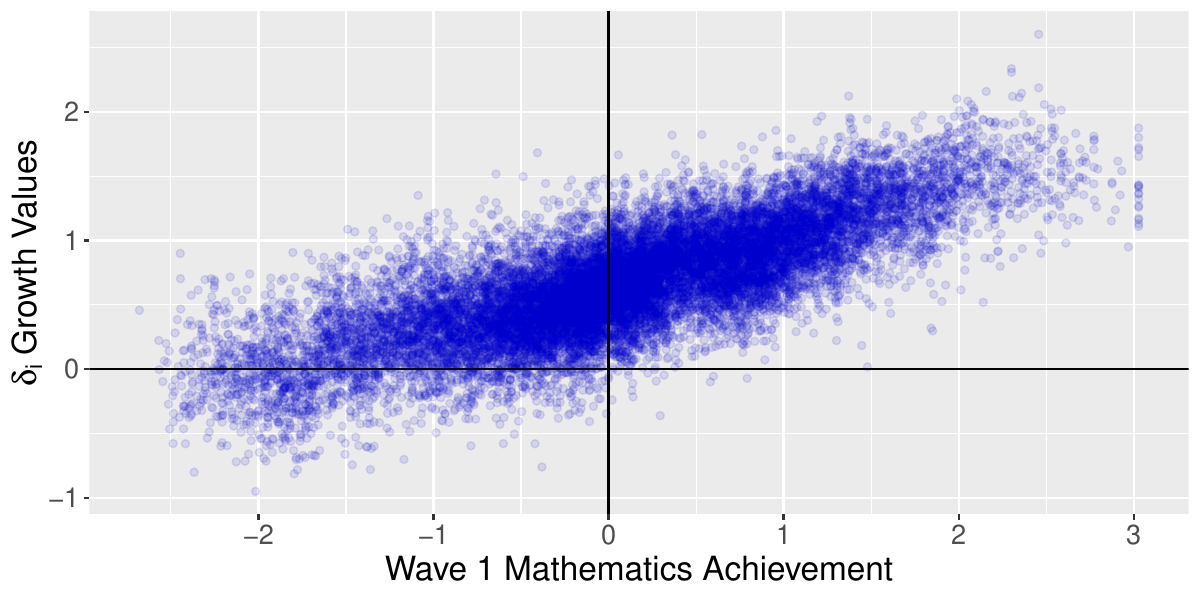}
    \caption{Scatterplot of the relationship between Wave 1 achievement and predicted $\delta_{i}$ values. Students with initially high levels of academic achievement are predicted to increase their achievement by higher amounts than their peers.}
    \label{ach_growth}
\end{figure}

To identify key moderating variables contributing to the variation in $\delta_{i}$ values, variable importance measures were calculated for the $\delta()$ trees by counting how often different variables were selected for the splitting rules used in this part of the model. This investigation identified the achievement of the students measured at Wave 1 as being highly influential. Prompted by this finding, we created Figure~\ref{ach_growth} which shows a scatter plot of the $\delta_{i}$ predictions versus the achievement of the students measured at Wave 1. The very strong positive relationship between Wave 1 achievement and predicted growth indicates that students who initially perform well in mathematics are predicted to increase their achievement by substantially more than those with lower achievement levels. At the extremely high levels of Wave 1 achievement, students are predicted to increase achievement by 1.5 units on average, while for students at the opposite end of the spectrum, growth in achievement is minimal. This observation points to a widening achievement gap between students at the high and low ends of the achievement spectrum \citep{mccall2006achievement, rowley2020contextualising}.

\begin{figure}
    \centering
    \includegraphics[width=15cm]{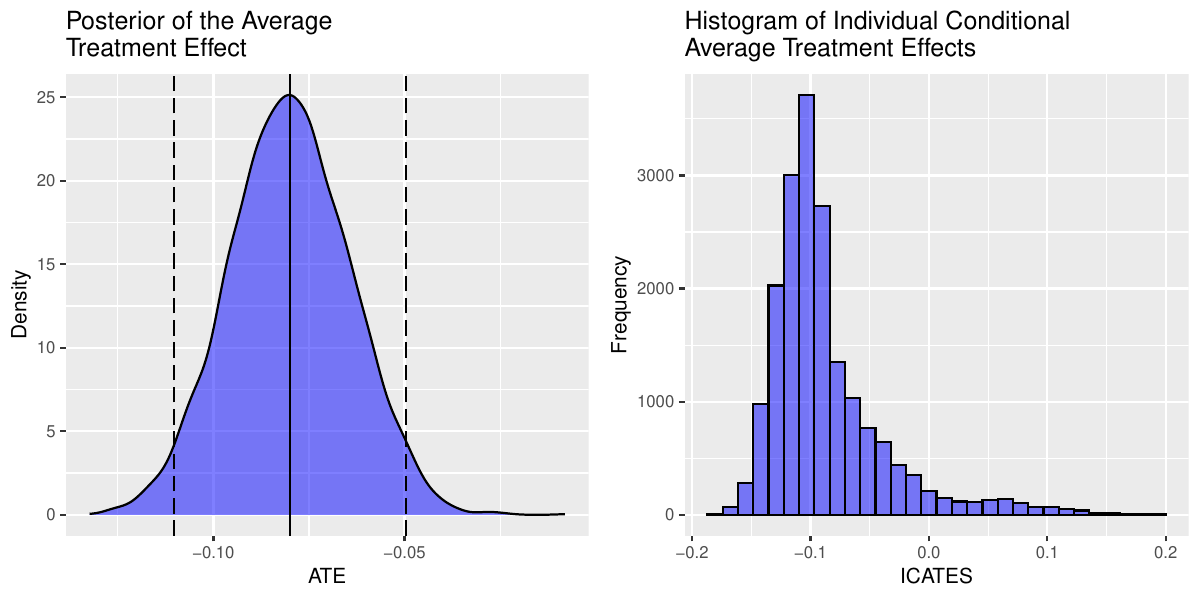}
    \caption{The posterior distribution for the Average Treatment Effect (ATE) is shown on the left, and a histogram of the individual conditional average treatment effects is provided on the right. The solid line shows the posterior mean, while the dashed lines indicate a 95\% credible interval. An interesting subgroup of students on the right tail of the histogram are predicted to benefit from part time work.}
    \label{ate_plot}
\end{figure}

The posterior distribution of the average treatment effect for working part-time at an intensity of greater than 20 hours per week between Waves 1 and 2 is displayed in Figure~\ref{ate_plot}. The posterior mean of the ATE is approximately -0.08, with a 95\% credible interval ranging from -0.050 to -0.110. This indicates that on average, part-time work is expected to reduce the growth in student achievement between Waves 1 and 2 by between 0.050 and 0.110 units. To contextualise this effect size, note that the standard deviation of the $\delta_{i}$ growth values in achievement is approximately 0.44. Thus, the observed effect size corresponds to a decrease in achievement growth by nearly 0.2 standard deviations, which can be considered a medium to large effect size \citep{kraft2020interpreting}.

\begin{figure}
    \centering
    \includegraphics[width=15cm]{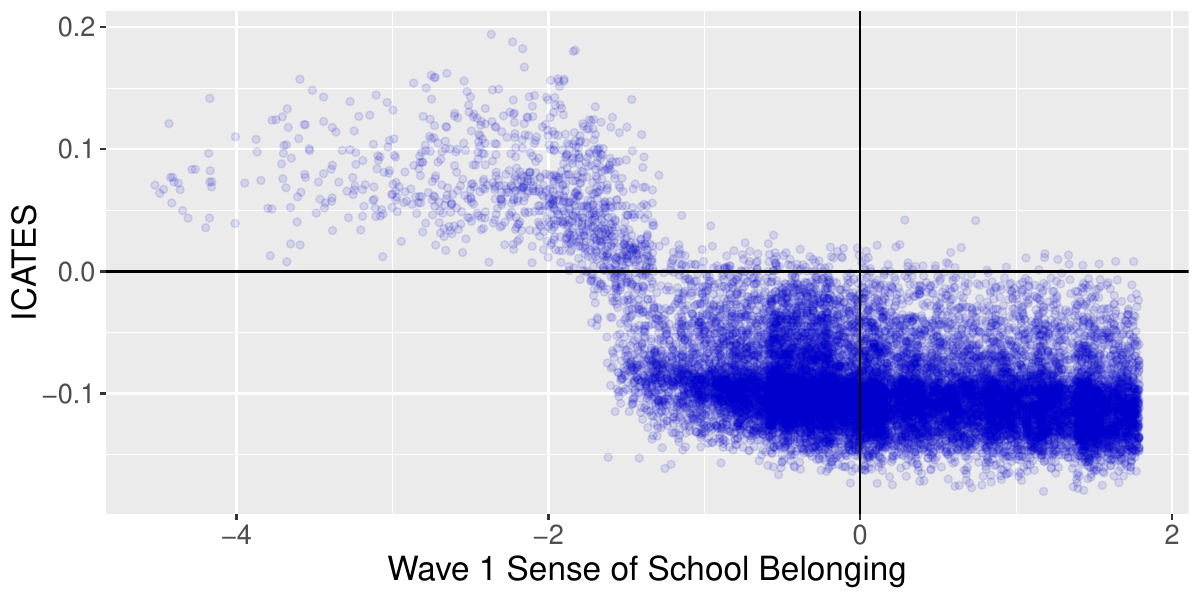}
    \caption{Scatterplot of the relationship between Wave 1 school belonging and the effect of working part-time. The effect of part-time work is negative for most students, but for a subgroup of students with low sense of school belonging the predicted effect is positive.}
    \label{belonging_icates}
\end{figure}

A histogram of the individual conditional average treatment effects (ICATEs) for each of the students in the sample can be found in Figure~\ref{ate_plot}. The majority of the ICATEs are centered quite close to the ATE of -0.08, but there are also signs of heterogeneity. Notably, there is an interesting tail of the histogram stretching across into a positive area where the effect of part-time work is actually predicted to have a positive effect on achievement growth. To explore this finding further, we calculated variable importance metrics for the $\tau()$ trees in our model to identify any variables that might strongly moderate the treatment effect. The most influential variable resulting from this analysis was a measure of the students' sense of school belonging at Wave 1 of the study. Figure~\ref{belonging_icates} visualises this variable's relationship with the ICATEs from the model. The results suggest that the students predicted to experience a positive effect from part-time work are those with an initially low sense of school belonging.

This interesting finding, which might initially appear quite strange, aligns well with some `traditional' views that part-time work can benefit students. Early research has suggested, for example, that part-time employment can provide students with greater time management skills \citep{robotham2012student}, and other benefits such as a sense of purpose and responsibility. These benefits can be especially pronounced among students with low achievement or a diminished sense of belonging in school \citep{king1989improving, steinberg1982high}. This sense of purpose and responsibility acquired through part-time work could serve to re-focus students, leading to spillover effects benefiting their academic performance \citep{zimmerman2005homework}. Therefore, while part-time work may be associated with negative outcomes for the majority of students, there may be certain subgroups, such as students experiencing a low sense of belonging in school, who may experience positive effects from employment.

In summary, this section presented two key findings from the analysis of the HSLS data. Firstly, substantial variation was observed in the extent to which students improved their achievement between Waves 1 and 2. Further analysis showed that this variation was driven primarily by the baseline achievement levels of the students, with initially high performing students showing much higher growth than their peers. These students, starting from a solid foundation of high achievement may find it easier to build upon their academic progress, as they are in a better place to acquire and digest new knowledge in class. The second key finding was that on average, part-time work had a modest, but negative effect on the growth of student achievement. This supports the ``zero-sum" argument that part-time work detracts from study time, homework completion, and rest, hindering academic progress as a result. A notable exception was that students with initially low school belonging might actually benefit from part-time work, highlighting the ability of our model to capture complex relationships between student performance and employment.

\section{Discussion}
\label{discussion}

Drawing on longitudinal data from the High School Longitudinal Study of 2009, our study introduced an innovative method for modeling growth in student achievement. Our model also estimates the causal impact of interventions such as part-time work on this growth. By extending Bayesian Additive Regression Trees \citep{chipman2010bart} and Bayesian Causal Forests \citep{hahn2020bayesian}, the primary strength of our model lies in its ability to flexibly capture both individual growth trajectories in student achievement and the potentially heterogeneous treatment effects of part-time work, which may be influenced by various covariates. This approach contrasts with many existing methods that either lack the flexibility to model individual variations or are confined to single time-point observational data, precluding an analysis of achievement growth over time.

Our model was also equipped with two special features that allowed it to handle missing data in the covariates and the treatment status indicator. Simulation study results from Section~\ref{sim section} provide strong support for the impressive predictive performance of the model, which demonstrated clear advantages over three competing methods when tasked with predicting growth values at the individual student level, and heterogeneous treatment effects. Close to ideal coverage rates were also achieved. The proposed model also showed strong performance in a second simulation study, matching the performance of two correctly specified models designed specifically for use with longitudinal datasets.

The results from our model application to the motivating HSLS data produced some interesting findings. First, the model was able to reveal a large disparity in the predicted growth values among students with initially high and low levels of academic achievement. This finding of a widening achievement gap underscores the importance of early interventions in schools and academic institutions. By addressing achievement gaps at the elementary and middle school levels, policy decisions can prevent these disparities from becoming entrenched. This is especially important given previous research which indicates that it becomes much more challenging to effectively remedy these gaps by the ninth or eleventh grade \citep{morgan2016science}.

On average, part-time work was found to have a negative effect on student achievement, with the 95\% credible interval for the ATE ranging from -0.050 to -0.110. This is important, as we calculated nearly 50\% of students in our sample participated in some level of part-time work during high school, and more than 15\% of students participated in intensive part-time work, requiring upwards of 20 hours of work a week. Large amounts of heterogeneity were apparent in the ICATEs, however, and an analysis of the variable importance metrics from the model identified sense of school belonging during Wave 1 as a significant contributor to this variation. The finding that students with a low sense of school belonging may actually be benefiting slightly from part-time work ties in with previous findings that show students can benefit from the routine, sense of purpose and responsibility that part-time work can provide \citep{robotham2012student, king1989improving, steinberg1982high}. From a policy perspective, however, we do not recommend that students beginning to disengage from the school system should take on intensive part-time work. Instead, we suggest that further research is needed to explore how disengaging students can be encouraged to find a sense of purpose or routine through other activities such as sports or youth programs. Alternatively, part-time work with moderate hours may be a more balanced approach.

A limitation of the model proposed in our study is that owing to the fact each growth period and associated treatment effect is dedicated a separate BART model, the computational cost of running the model may become quite large in settings with many waves of data. Replacing the BART models with more efficient XBART models as in \citet{he2023stochastic} and \citet{krantsevich2023stochastic} would therefore make a promising area for future work, widening the applicability of the proposed method. 

Given the flexibility and widely adopted nature of the underlying BART framework, a natural extension of the longitudinal causal model adopted in our study might be to survival data \citep{sparapani2016nonparametric}. Other natural extensions could include allowing multivariate \citep{mcjames2024bayesian} or multinomial outcomes \citep{murray2021log}, or the incorporation of random effects \citep{wundervald2022hierarchical, yeager2022synergistic}. Additionally, given the specificity of our results to a representative sample of ninth to eleventh grade high school students from the US, an application of a similar model to other countries or grade levels would be of interest. More generally, we expect that the model's flexibility will allow it to be applied to a wide variety of datasets across diverse fields and application areas.

\spacingset{1}

\newpage
\section*{Supplementary Materials - Table of Summary Statistics}
\setcounter{page}{1}

\begin{table}[h!]
\spacingset{1}
\centering
\rotatebox{90}{
\resizebox{1.2\textwidth}{!}{\begin{tabular}{l|ccc}
    \toprule
    \textbf{Variable} & \textbf{Proportion Wave 1} & \textbf{Achievement Wave 1} & \textbf{Achievement Wave 2} \\
    \midrule
    \textbf{Student Gender} & & &  \\
    \hspace{0.5cm} Male & 50.3\% & -0.08 & 0.63 \\
    \hspace{0.5cm} Female & 49.7\% & -0.06 & 0.60 \\
    \midrule
    \textbf{First Language} & & &  \\
    \hspace{0.5cm} English Only & 82.3\% & -0.05 & 0.63  \\
    \hspace{0.5cm} English and Other & 6.1\% & -0.15 & 0.58 \\
    \hspace{0.5cm} Other & 11.5\% & -0.16 & 0.58 \\
    \midrule
    \textbf{Family Setup} & & &  \\
    \hspace{0.5cm} Live With Both Biological Parents & 43.2\% & 0.21 & 0.93  \\
    \hspace{0.5cm} Other Arrangement & 32.9\% & 0.02 & 0.71 \\
    \hspace{0.5cm} No Response & 23.9\% & -0.38 & 0.25 \\
    \midrule
    \textbf{Parent Education Level} & & & \\
    \hspace{0.5cm} Less Than Bachelor's Degree & 47.7\% & -0.23 & 0.39  \\
    \hspace{0.5cm} Bachelor's Degree or Higher & 28.4\% & 0.45 & 1.23 \\
    \hspace{0.5cm} No Response & 23.9\% & -0.38 & 0.25 \\
    \midrule
    \textbf{Student Future Expectations} & & & \\
    \hspace{0.5cm} Less Than Bachelor's Degree & 21.5\% & -0.56 & 0.05  \\
    \hspace{0.5cm} Bachelor's Degree or Higher & 56.8\% & 0.18 & 0.89 \\
    \hspace{0.5cm} Student Not Sure & 21.7\% & -0.25 & 0.41 \\
    \midrule
    \textbf{School Type} & & &  \\
    \hspace{0.5cm} Public & 92.8\% & -0.11 & 0.57  \\
    \hspace{0.5cm} Catholic or Private & 7.2\% & 0.35 & 1.21 \\
    \bottomrule
\end{tabular}}
\captionsetup{width=1\textwidth}}
\caption{Summary statistics for selected categorical variables. The proportion column provides the proportion of students belonging to each category in Wave 1, while the Wave 1 and Wave 2 achievement columns provide the mean achievement level within each group.}
\label{side_table}
\end{table}

\newpage
\section*{Supplementary Materials - LBCF Diagram}

\begin{figure}[htp]
    \centering
    \includegraphics[width=\linewidth]{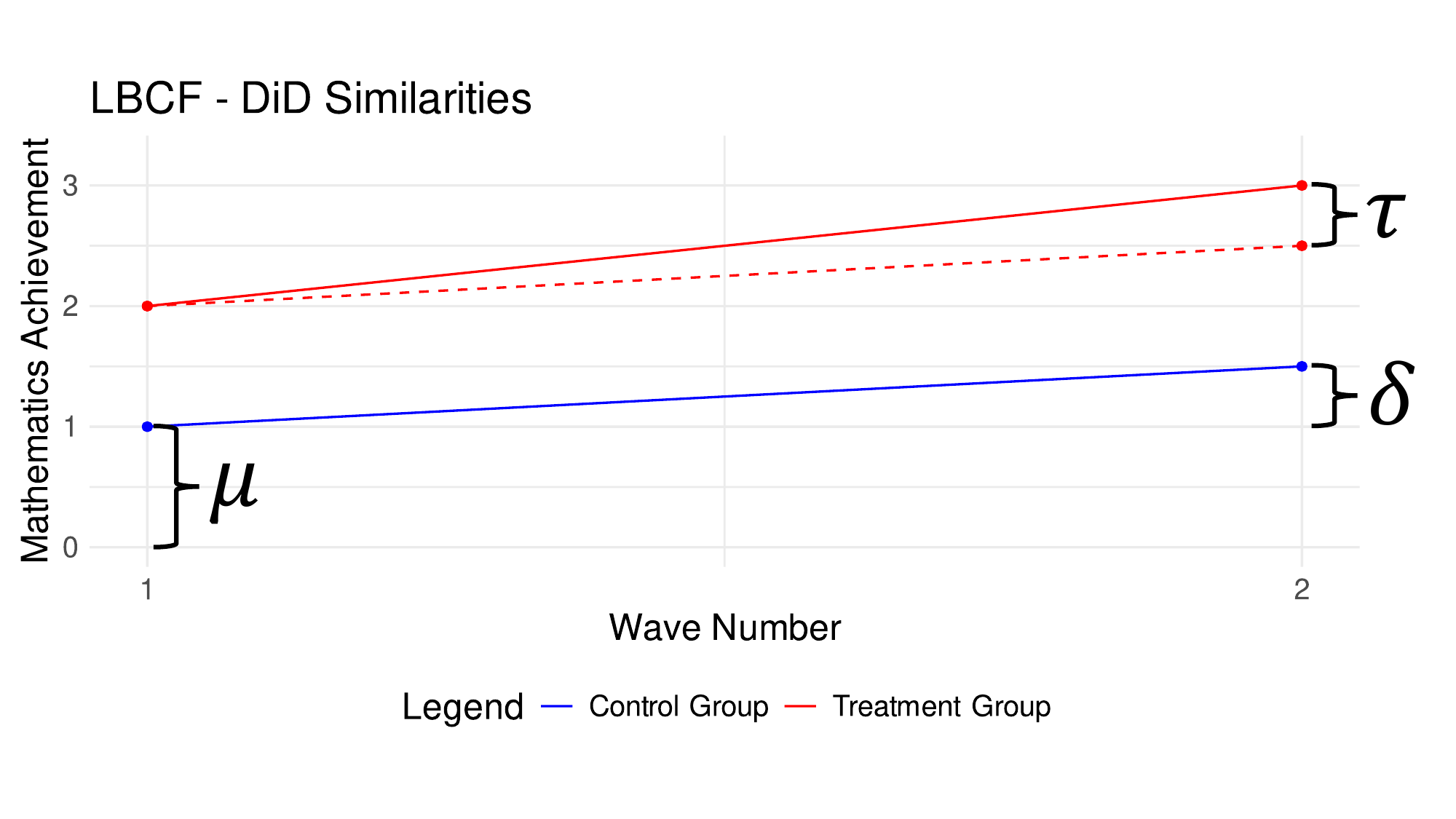}
    \caption{Diagram of how the proposed LBCF model fits into the framework of the difference-in-differences approach. Two observations are shown for the purposes of illustration - one from an imaginary control group, and one from a corresponding treatment group. The solid lines indicate the realised achievement trajectories, while the dashed line in red indicates a counterfactual trajectory for the treated unit had it actually not received treatment. Initial achievement estimates at Wave 1 are provided by $\mu$. The expected growth (difference) in achievement without treatment is provided by $\delta$, while the effect of treatment on this growth (the difference-in-differences) is captured by $\tau$. Note that while only one $\mu$ value is indicated in the diagram to avoid overprinting, the model does in fact provide individual $\mu$ estimates for every observation. Similarly, individual estimates are provided for each of the $\delta$ and $\tau$ estimates as well.}
\end{figure}

\newpage
\section*{Supplementary Materials - LBCF Algorithm}

\footnotesize
\RestyleAlgo{boxruled}
\begin{algorithm}[H]
 \KwData{Outcome variable $y_{i,t}$ (response for individual $i$ at time $t$ of $T$ time periods)\;
        time varying covariates $x_{i,t}$ (covariates collected on individual $i$ up to time $t$)\;
        treatment variable $Z_{i,t+1}$ (to indicate if individual $i$ received treatment between periods $t$ and $t+1$: 1 for treatment, 0 for control)}
 \KwResult{Posterior list of trees, values of $\sigma^2$, fitted values $\hat{\mu}_{i}$, $\hat{\delta}_{i,t}$, and $\hat{\tau}_{i,t}$}
 \textbf{Initialisation\;}
 	Hyper-parameter values of $\alpha_{\mu}$, $\beta_{\mu}$, $\alpha_{\delta}$, $\beta_{\delta}$, $\alpha_{\tau}$, $\beta_{\tau}$, $\sigma_{\mu}^2$, $\sigma_{\delta}^2$, $\sigma_{\tau}^2$, $\nu$, $\lambda$\;
	Number of $\mu$ trees $n_{\mu}$\;
        Number of $\delta$ trees $n_{\delta}$\;
        Number of $\tau$ trees $n_{\tau}$\;
	Number of iterations $N$\;
	Initial value $\sigma^2 = 1$\;
	Set $\mu$ trees $T_j$; $j=1, \ldots, n_{\mu}$ to stumps\;
	Set $\delta$ trees $T_j$; $j=1, \ldots, n_{\delta}$ to stumps\;
        Set $\tau$ trees $T_j$; $j=1, \ldots, n_{\tau}$ to stumps\;
	Set terminal node parameters of all $\mu$, $\delta$, and $\tau$ trees to 0\;
\For{iterations $i$ from 1 to $N$}{
	\For{$\mu$ trees $j$ from 1 to $n_{\mu}$}{
		Compute partial residuals from $y$ minus predictions of all trees except $\mu$ tree $j$\;
		Grow a new tree $T_j^{new}$ based on grow/prune/change/swap\;
		Accept/Reject tree structure with Metropolis-Hastings step using $P(T_{\mu,j}|R_{\mu,j}, \sigma^2)\propto P(T_{\mu,j})P(R_{\mu,j}|T_{\mu,j}, \sigma^2)$\;
		Sample $\mu$ values from normal distribution using $P(M_{\mu,j}|T_{\mu,j}, R_{\mu,j}, \sigma^2)$\;
	}
        \For{Time periods t from 2 to T}{
	\For{$\delta_{t}$ trees $j$ from 1 to $n_{\delta}$}{
		Compute partial residuals from $y$ minus predictions of all trees except $\delta_{t}$ tree $j$\;
		Grow a new tree $T_j^{new}$ based on grow/prune/change/swap\;
		Accept/Reject tree structure with Metropolis-Hastings step using $P(T_{\delta_{t},j}|R_{\delta_{t},j}, \sigma^2)\propto P(T_{\delta_{t},j})P(R_{\delta_{t},j}|T_{\delta_{t},j}, \sigma^2)$\;
		Sample $\delta_{t}$ values from normal distribution using $P(M_{\delta_{t},j}|T_{\delta_{t},j}, R_{\delta_{t},j}, \sigma^2)$\;
	}
        \For{$\tau_{t}$ trees $j$ from 1 to $n_{\tau}$}{
		Compute partial residuals from $y$ minus predictions of all trees except $\tau_{t}$ tree $j$\;
		Grow a new tree $T_j^{new}$ based on grow/prune/change/swap\;
		Accept/Reject tree structure with Metropolis-Hastings step using $P(T_{\tau_{t},j}|R_{\tau_{t},j}, \sigma^2)\propto P(T_{\tau_{t},j})P(R_{\tau_{t},j}|T_{\tau_{t},j}, \sigma^2)$\;
		Sample $\tau_{t}$ values from normal distribution using $P(M_{\tau_{t},j}|T_{\tau_{t},j}, R_{\tau_{t},j}, \sigma^2)$\;
	}
        }
	
	Get predictions $\hat{y}$ from all trees\;
	Update $\sigma^2$ with Inverse-Gamma distribution using $P(\sigma^2 | \hat{y})$\;
}
\caption{LBCF MCMC Algorithm}
\end{algorithm}

\end{document}